\documentclass[10pt,twocolumn,letterpaper]{article}

\usepackage{iccv}
\usepackage{times}
\usepackage{epsfig}
\usepackage{graphicx}
\usepackage{amsmath}
\usepackage{amssymb}
\usepackage[accsupp]{axessibility}
\usepackage{titling}
\usepackage{blindtext}
\usepackage{soul}
\newcommand{\suptit}{\supple}
\usepackage[breaklinks=true,bookmarks=false]{hyperref}
\usepackage[capitalise]{cleveref}

\iccvfinalcopy %

\begin{document}

\title{Generative Models for Multi-Illumination Color Constancy}

\author{Partha Das$^{1,2}$\\
{\tt\small p.das@uva.nl}
\and Yang Liu$^2$\\
{\tt\small yang@3duniversum.com}
\and Sezer Karaoglu$^{1,2}$\\
{\tt\small s.karaoglu@3duniversum.com}
\and Theo Gevers$^{1,2}$\\
{\tt\small th.gevers@uva.nl}
\and University of Amsterdam, The Netherlands$^1$
\and 3DUniversum, Amsterdam, The Netherlands$^2$
}
\date{}
\maketitle

\begin{abstract}
   In this paper, the aim is multi-illumination color constancy. However, most of the existing color constancy methods are designed for single light sources. Furthermore, datasets for learning multiple illumination color constancy are largely missing. We propose a seed (physics driven) based multi-illumination color constancy method. GANs are exploited to model the illumination estimation problem as an image-to-image domain translation problem. Additionally, a novel multi-illumination data augmentation method is proposed. Experiments on single and multi-illumination datasets show that our methods outperform sota methods.
\end{abstract}

\section{Introduction}

The color of an object is influenced by the illumination color. The process of recovering the original object color, independent of the illumination color, is called Color constancy. Previous approaches solve the problem by enforcing various priors~\cite{Land1977, Buchsbaum1980, Qian2019}. More recently, with the availability of relatively large scale datasets~\cite{Ciurea2003,Finlayson2017,Banic2018,Aytekin2017}, learning based approaches are explored~\cite{Lou2015,Bianco2008}. Various learning based approaches are proposed such as local-global patch correction~\cite{Gijsenij2012-1} and exemplar patch based correction~\cite{Joze2014}, that show robust performance. However, these methods predict the illumination vector directly. This limits the learning capability of the network by limiting all the pixels to be supervised by the illumination vector. Modelling the problem as an image-to-image translation could provide a solution to this, allowing finer pixel wise supervision. However, standard encoder-decoder network works by minimising the Euclidean distance between the prediction and the ground truth. This may lead to blurry outputs and border artefacts, while being fully dependent only on the input.

Generative Adversarial Networks~\cite{Goodfellow2014} (GAN) solve this problem by learning an image-to-image translation as a domain transformation. In this setting, the generator learns an expanded pixel-wise supervision, while the discriminator allows the network to learn domain specific information. To exploit the unified framework, we propose to use GANs to model color constancy as a domain transformation, instead of image to illumination prediction.

However, the previous methods are often limited to a single illumination assumption. For multi-illumination scenes, each pixel in an image is influenced by different light sources. GAN models like Pix2Pix or CycleGAN learn a global transformation. Even with pixel wise supervision, Pix2Pix learns a pixel wise transformation, excluding the spatial context. In contrast, we propose a hybrid learning and seed based multi-illumination color constancy. Using physics based (initial) seed points, the network is provided with illumination regions. The network then diffuses these seed points over the image for a multi-illumination prediction. To model mixtures of lights at a pixel, an illumination probability is predicted. This allows the model to compute per-pixel mixture probability belonging to the different illuminations. Finally, due to the lack of multi-illumination datasets, a data augmentation method is proposed to create a new multi-illumination dataset. In summary, the proposed contributions of this work are as follows: 

\begin{itemize}
    \item We propose to model the problem of color constancy as a domain transformation with GANs.
    \item We propose a new method for exploiting physics based initial seed points for multi-illumination color constancy.
    \item We propose a novel data augmentation method to create multi-illumination dataset using off-the-shelf methods and single illumination datasets.
\end{itemize}

\section{Related Work}

Color constancy methods can be classified into optimization-based methods and learning based methods.

\textbf{Optimization based methods:} These methods exploit various low-level image statistics or priors ~\cite{Land1977, Buchsbaum1980, Jepson1987, Hussain2018, Finlayson2004, Weijer2007}. New features are proposed in~\cite{Qian2019} tackling both single and multi-illumination in a learning free manner. Obviously, these methods may fail when the underlying assumptions/priors are violated.

\textbf{Learning based methods:} Various learning based methods are proposed such as~\cite{Rosenberg2004,Funt2004,Bianco2010,Gijsenij2010,Beigpour2014,Cheng2015} using classical learning methods like Bayesian frameworks, Support Vector Machines, Decision Forests, and Conditional Random Fields. Exemplar based methods~\cite{Joze2014} explore multi-illumination settings using pixel independent features. With the availability of larger datasets, CNNs are proposed~\cite{Lou2015,Bianco2015,Oh2017,Hu2017}. These methods employ deep learning frameworks in an end-to-end manner to model the problem as an illumination prediction. While these previous works focused on a single image,~\cite{Qian2017} integrates an implicit illumination constraint by using multiple frames and a deep recurrent network.~\cite{Abdelhamed2021}, on the other hand, uses a two-camera-one-subject setup to estimate the illumination. Finally,~\cite{Yan2018, Sidorov2018} formulated the problem into an image-to-image transformation. However, the methods suffer from dataset illumination bias.

In contrast to previous work, this paper proposes to approach the problem of color constancy as a domain transformation. Following our work~\cite{Das2018}, different GANs are first studied for the problem of single illumination color constancy. Their shortcomings for a multi-illumination setting are analyzed. These observations are then employed to propose an end-to-end hybrid multi-illumination color correction pipeline.

\section{Methodology}

\subsection{Color Constancy}

The problem of color constancy can be defined as a process of chromatic adaptation~\cite{Fairchild2013}:
\begin{equation} \label{eq:canon}
    I = W * L\;,    
\end{equation}

\noindent where, $W$ is the image under a canonical white illumination and $L$ is the unknown colored illumination. The $*$ operation denotes the pixel-wise multiplication between the components. Thus, $L$ acts as a scaling term on $W$ to generate the final $I$. All the components are in linear $RGB$. For a single light source, L is uniformly constant, while for multi-illumination L is varied.

The process of color correction can thus be approximated as modifying the gains of the $RGB$ channels independently. To recover the color corrected image ($W$), the Von Kries~\cite{Kries1970} method can be used as follows:

\begin{equation} \label{eq:vonKries}
\begin{aligned}
\begin{pmatrix}
R_{c} \\
G_{c} \\
B_{c}
\end{pmatrix} =
\begin{pmatrix}
L_{R} & 0 & 0 \\
0 & L_{G} & 0 \\
0 & 0 & L_{B}
\end{pmatrix}
\begin{pmatrix}
R_{u} \\
G_{u} \\
B_{u}
\end{pmatrix}\;,
\end{aligned}
\end{equation}

\noindent where $R_u$, $G_u$, $B_u$ denote $RGB$ channels of an image under an unknown colored illumination source, $R_c$, $G_c$, $B_c$ represent $RGB$ channels under a canonical (corrected) illumination, and $L_R$, $L_G$ and $L_B$ correspond to the unknown illumination color. 

Traditional approaches directly recover the illumination color from the input image. This forces the network to use the same supervision and ``copy'' it over all the pixels. In contrast, it is proposed to model the process of color constancy as domain mapping, i.e. from color biased ($I$) image to color corrected ($W$) image. This allows the network to be flexible to both single illumination and multi-illumination cases. GANs are used for this image-to-image translation, due to their ability to learn effective domain translation.

\subsubsection{Single illumination Color Constancy}

In this section, three GAN models are adapted for the problem of color constancy under a single unknown illumination. The color biased image is denoted as $I$, while the color corrected image is denoted as $W$. The entire domain of color biased images is denoted by $X$ and color corrected image domain as $Y$, that is $I \in X$ and $W \in Y$.

\textbf{Pix2Pix~\cite{Isola2017}:} Pix2pix employs a U-net architecture~\cite{Ronneberger2015} as the generator. Skip connections~\cite{Mao2016} allow the net to transfer high frequency details. $PatchGAN$ is introduced as the discriminator that enforces high frequency consistency through a Markov random field. Standard GAN losses and L1 losses on paired cross domain images are used to train the network. This allows the network to learn a pixel-wise transfer between domains while keeping the underlying image structure consistent.

This formulation is particularly well suited to the problem of color constancy. The problem is defined as a domain transformation of the image taken under an unknown illumination ($I$) to the color corrected image ($W$). The L1 loss makes the network approximate a color correction function, while the adversarial loss ensures that the transformation is close to the target domain. This allows the network to predict color corrected images directly, while removing the unknown illumination color influence. Unfortunately, the L1 loss necessitates dense ground-truth pairs.

\textbf{CycleGAN~\cite{Zhu2017}:} CycleGAN improves on Pix2Pix by introducing a forward and backward cycle. The network learns a closed cycle through: 1) forward transformation ($I \rightarrow W$), and 2) backward transformation ($W \rightarrow I$). This allows the network to be trained unsupervised. The supervision signal is obtained from the outputs of the forward and backward cycles, called the cycle consistency loss. Given two generators, G (the forward generator) and F (the backward generator), we have:

\begin{equation}\label{eq:cycle}
    \mathcal{L}_{cycle}(G,F) = E_{x \sim p_{data}(x)}\;||F(G(x)) - x||_{1}^{1}\;,
\end{equation}

The network then learns the cycles $G(X) \rightarrow Y'$ and $F(Y') \rightarrow X'$. This leads to learning a global transformation to approximate the color correction. Due to the unsupervised nature of training, any color biased and color corrected image can be assigned to the respective domain without the need for a dense ground-truth correspondence to train the network. Unfortunately, the network can learn only a single type of illumination correction for the color biased domain.

\textbf{StarGAN~\cite{Choi2018}:} To handle multi-domain translations simultaneously, StarGAN introduces an additional conditioning on the domain. The base model architecture of CycleGAN is used, with a domain vector on top of it. This vector specifies which target domain the network should map the input to. Consequently, the discriminator learns to classify the domain, in the form of an additional output. This allows the network to adversarially learn specific domain transformations conditioned on the input vector. Both the forward and backward cycles from CycleGAN are preserved, allowing the network to be trained in an unsupervised manner.

The ability to learn a transformation between various domains translates well to the problem of color constancy. This allows the network to learn a common transformation between different illumination types using a single model. Furthermore, the common domain defined transformation, yields a network having a consistency check in the form of whether the model can recover the same image from different illumination colors. The domains are also learned in an unsupervised manner, suited to train with limited data.

\textbf{Limitations:} The different GANs provide useful formulations for the problem of color constancy. However, (1) the networks are merely suited to the problem of single illumination color constancy, (2) the networks are not tailored to model the color correction process, (3) the networks implicitly learn the illumination as a latent state, and (4) the transformations learned are often global and are lacking spatial context. As a result, these networks needs to be adapted to multi-illumination color constancy.

\subsubsection{Multi-Illumination Color Constancy}
\begin{figure}[h]
    \centering
    \includegraphics[width=\linewidth]{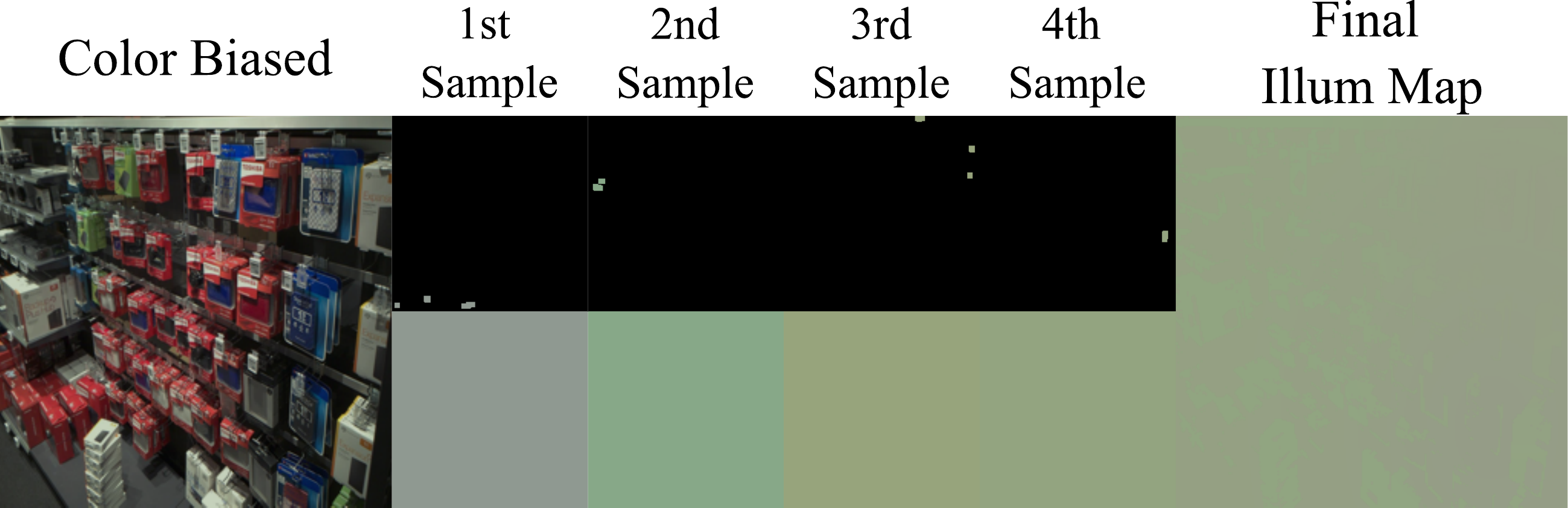}
    \caption{Examples of the dataset augmentation. The 1st-4th samples are the random samples from the respective illuminants. The color patches below it are the illumination themselves. The Final Illum Map is the total illumination maps consisting of the four illuminants. All inputs are of the same spatial resolution as the color biased image. They have been resized here for visualisation.}
    \label{fig:inputs}
\end{figure}

In this section, a sample-based learning approach to solve the multi-illumination color constancy is proposed. In this, no assumption on the number or spatial influence of the illuminants are made. Instead, random initial seed points are provided to the network as inputs. The ground truth illumination map is used to randomly sample seed points for different light sources during training. The network uses these initial points to disentangle the illumination influence for all pixels. However, this might lead the network to be biased to the initially selected seeds. To avoid the bias, the color biased image ($I$) is also provided to the network as an additional input. The color biased image provides the network with a global cue, while the initial seed points provide a local cue. Using these cues, the network can learn to diffuse the initial sample points through the image and in-paint an illumination map. The input modalities are given in figure~\ref{fig:inputs}.

For multi-illumination, every pixel can have varying influence from the different illuminants. To model this, the network predicts a probability map, instead of the color corrected image. In this, each pixel is assigned a probability of belonging to a specific illumination. This transforms the problem to a per-pixel classification task, rather than a global transformation. As a result, the network does not have to cluster the illuminants in the latent space, while having a realistic illumination mixture for each pixel. The output of the network is a $N$-channel probability map for $N$ illuminants. Each of the channels represents the probability of a single pixel belonging to the $N$-th illumination. The predicted illumination map is reconstructed using the following:

\begin{equation}
    \label{eq:illum_pred}
    L_{pred} = \sum_{i=1}^N P_i \times I_i
\end{equation}

\noindent where, $P_i$ is the $i$-th channel in the $N$ channel probability map predicted by the network and $I_i$ is the corresponding $i$-th illumination. $L_{pred}$ is the predicted illumination map obtained from the probability map.

\begin{figure}
    \centering
    \includegraphics[width=\linewidth]{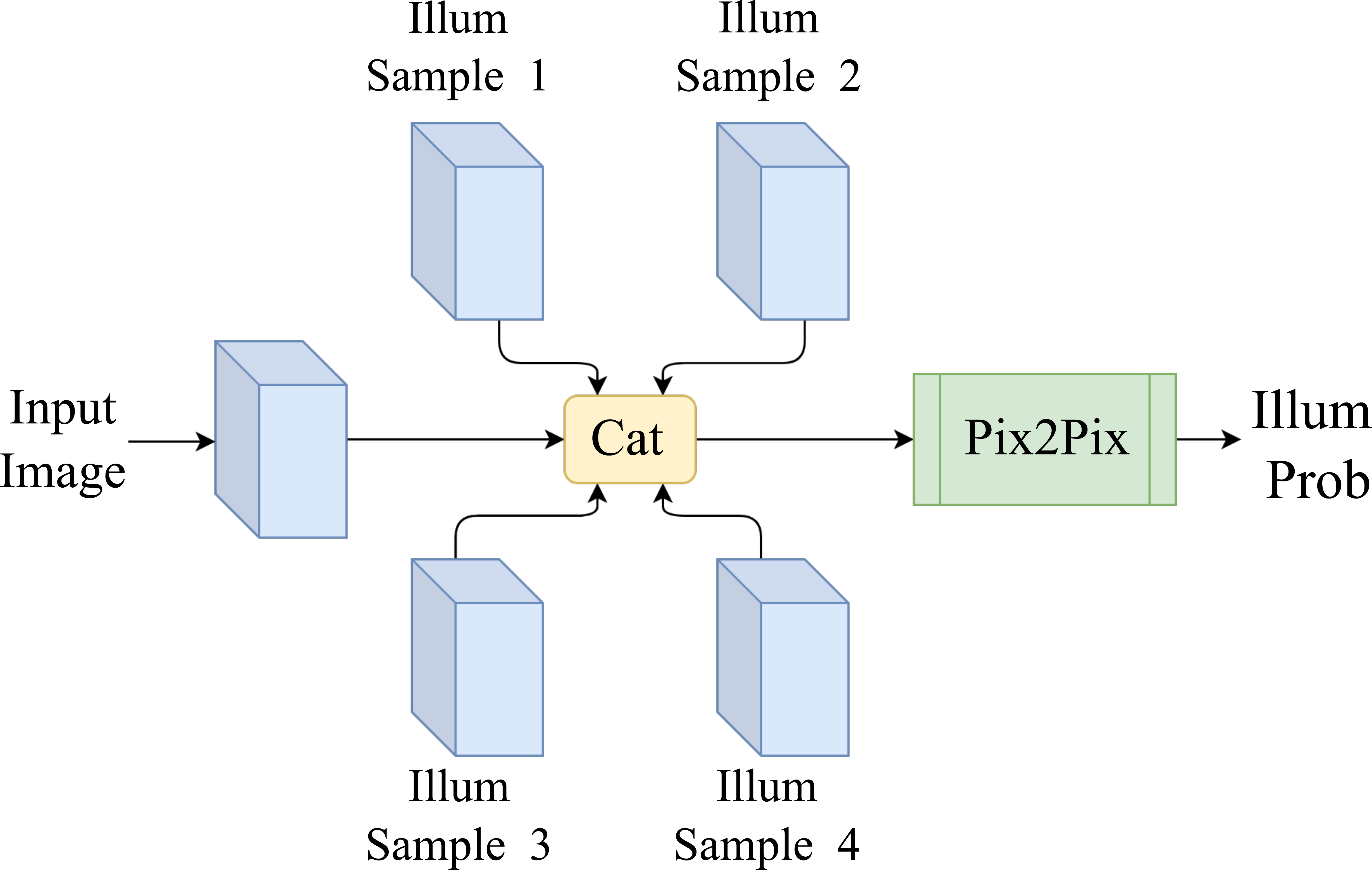}
    \caption{Overview of the proposed multi-illumination color constancy network. The samples are a concatenation of the image samples and the corresponding masks. The cube denotes convolutions. The output of the network is a probability map, which is processed using equation~\eqref{eq:illum_pred} to obtain the illumination map. Please refer to the supplementary material for details.}
    \label{fig:gen_overview}
\end{figure}

Finally, to let the network learn a realistic transformation, a PatchGAN discriminator is used. This helps to keep the network from learning a dataset bias. An overview of the generator is shown in figure~\ref{fig:gen_overview}.

During inference, Grayness Index~\cite{Qian2019} is used to determine the illumination clusters and randomly sample from those clusters. The grayness index is a physics-based, non-learning estimation. Pix2Pix is used as the base model architecture since it is more suited for local illumination estimation which is beneficial for multi-illumination color constancy. The direct supervision in Pix2Pix is also useful for the probability output of the network. In CycleGAN, the transformation would have to be done from image and sampled seed to probability, in the forward cycle, and then from probability to image and sampled seed for the backward cycle. This cycle consistency would be more difficult for a GAN to learn since the probability map from the forward cycle would not give enough information for the backward cycle. Additionally, compared to CycleGAN and StarGAN, Pix2Pix consists of just a single generator and discriminator, hence it has comparatively lower parameters, making it quicker and easier to train. The network also provides a local transformation, which is useful for scenes with multiple light sources since each pixel can be affected by the combined effect of all the light sources. To integrate the sample dispersion learning mechanism, an additional shallow network before the Pix2Pix generator is added. This consists of two layers of convolutions to condition the incoming features, increasing and then squeezing the feature channels. For the input illumination samples, the sample mask is also concatenated. This allows the network to learn a diffusion of the samples to their correct regions and back-propagate the errors from the seed sample.  Details on the shallow network can be found in the supplementary materials.

\textbf{Loss Functions:} The predicted probability map is constrained by a loss on the predicted illumination calculated using~\eqref{eq:illum_pred}:

\begin{equation}
    \label{eq:loss_illum}
    \mathcal{L}_{illum} = \frac{1}{M}\sum_{i=1}^M ||\hat{L_i} - L_i||
\end{equation}

\noindent where $M$ are the total number of pixels, $\hat{L_i}$ is the predicted illumination map and $L_i$ is the ground truth illumination map.

Furthermore, a loss on the color corrected image is added:

\begin{equation}
    \label{eq:loss_rgb}
    \mathcal{L}_{rgb} = \frac{1}{M}\sum_{i=1}^M ||\hat{W_i} - W_i||
\end{equation}

\noindent where $\hat{W_i}$ is the predicted color corrected image. This is obtained by using~\eqref{eq:vonKries} with the input color biased image and the predicted illumination map. The $W_i$ is the ground truth color corrected image.

Furthermore, to ensure that the probability map assigns the illuminants to the proper illumination region, an illumination mask loss is added. The initial seed points, given as input to the network, are compared to the corresponding points on the predicted illumination map.

\begin{equation}
    \label{eq:loss_masks}
    \mathcal{L}_{masks} = \sum_{j=1}^{N} ||\hat{L_j} * B - L_j * B||
\end{equation}

\noindent where $B$ are the binary masks denoting the locations of seed sampling points. This loss is calculated for each illumination.

Finally, we use an adversarial training to train the entire network:

\begin{equation}\label{eq:GANLoss}
    \begin{aligned}
            \mathcal{L}_{GAN} &= E[log D(I)]\;+\\
                                  &\;\;\;\;E[log(1 - D(G(\hat{I})))]\;,
    \end{aligned}
\end{equation}

\noindent where, $G(x)$ is the generator and $D(x)$ is the discriminator.

Thus, the total training objective for the network becomes:

\begin{equation}
    \label{eq:loss}
    \mathcal{L}_{total} = \mathcal{L}_{GAN} + \lambda (\mathcal{L}_{illum} + \mathcal{L}_{rgb} + \mathcal{L}_{masks})
\end{equation}

\noindent where, $\lambda$ is a hyper-parameter. This was empirically fixed to be $100$.

\section{Experiments}

\subsection{Datasets}

\subsubsection{Single Illumination Datasets}

The GAN models are trained and evaluated on 3 standard single illumination benchmarks: (1) The SFU Gray Ball~\cite{Ciurea2003} dataset, (2) the recalculated version of~\cite{Finlayson2017} called \emph{ColorChecker RECommended} (RCC) and (3) the Cube dataset~\cite{Banic2018}. For the experiments, the datasets are randomly split into 80\% training and 20\% testing. The reference objects in the images are masked out.

\subsubsection{Multi Illumination Datasets}

Large datasets for multi-illumination color constancy are difficult to obtain. Therefore, there are only a few datasets publicly available with limited number of images~\cite{Beigpour2014}. To train our network, a multi-illumination augmentation approach is proposed. Semantic segmentation algorithms are used to obtain object boundaries. Then, using a mixture model, the images are augmented with different illuminants. The illuminants are obtained from an existing multi-illumination dataset like MIMO~\cite{Beigpour2014}. The illuminants are then channel shuffled and combined with color corrected images using~\eqref{eq:vonKries} to obtain a multi illumination image. This approximates images taken from multi-illumination scenes with realistic illumination interaction, textures, and colors.

For the experiments, the Intel TUT~\cite{Aytekin2017} dataset, consisting of 5000 single illumination images, is used. Images are composed of various scenes, including indoors and outdoors. The dataset provides single illumination ground-truth for each image, recorded using a color checker object. To augment the dataset for multi-illumination, a semantic segmentation for each image is obtained, using Semantic Soft Segmentation~\cite{Aksoy2018}. The segmentation is then used to create the illumination map. For each illumination map obtained, random seed pixels from each of the illumination sections are selected. Thus, for each image, with $N$ illuminants, there are $N\times2+2$ image pairs: $N$ illumination color images, $N$ seed point masks, the multi-illumination color biased image and the corresponding color corrected image.

During inference, the grayness index~\cite{Qian2019} is used to obtain the initial multi-illumination clusters on the MIMO dataset. These clusters are then used to obtain the initial seed points, which are then passed on to the network. All numbers reported, and unless specifically mentioned, are from networks trained completely on the augmented Intel TUT dataset and tested on the MIMO dataset.

\subsection{Error Metrics}

The angular error between the ground-truth illumination $e$ and the estimated illumination $\hat{e}$ is reported: 

\begin{equation} \label{eq:angularError}
    d_{ang} = arccos(\dfrac{e \cdot \hat{e}}{||e|| \times ||\hat{e}||})\;,
\end{equation}

\noindent where $||.||$ is the L2 norm. For each image, the angular error is computed, and the mean, median, trimean, means of the lowest-error 25\%, the highest-error 25\%, and maximum angular errors are reported.

For the architectures that do not provide an illumination estimation directly, like the single illumination GAN experiments, the apparent illumination from the color corrected image is obtained. This is done by inverting equation~\eqref{eq:canon}, as follows:

\begin{equation} \label{eq:metric}
    e = I * \hat{W} ^{-1}\;,   
\end{equation}

\noindent where $I$ is the input image and $\hat{W}$ is the white balanced estimation from the network. Both $I$ and $\hat{W}$ are converted into the linear $RGB$ space before computing the estimated illumination.

\subsection{Single Illumination Color Constancy}

\subsubsection{GAN Comparison}

\begin{figure}
    \centering
    \includegraphics[width=\linewidth]{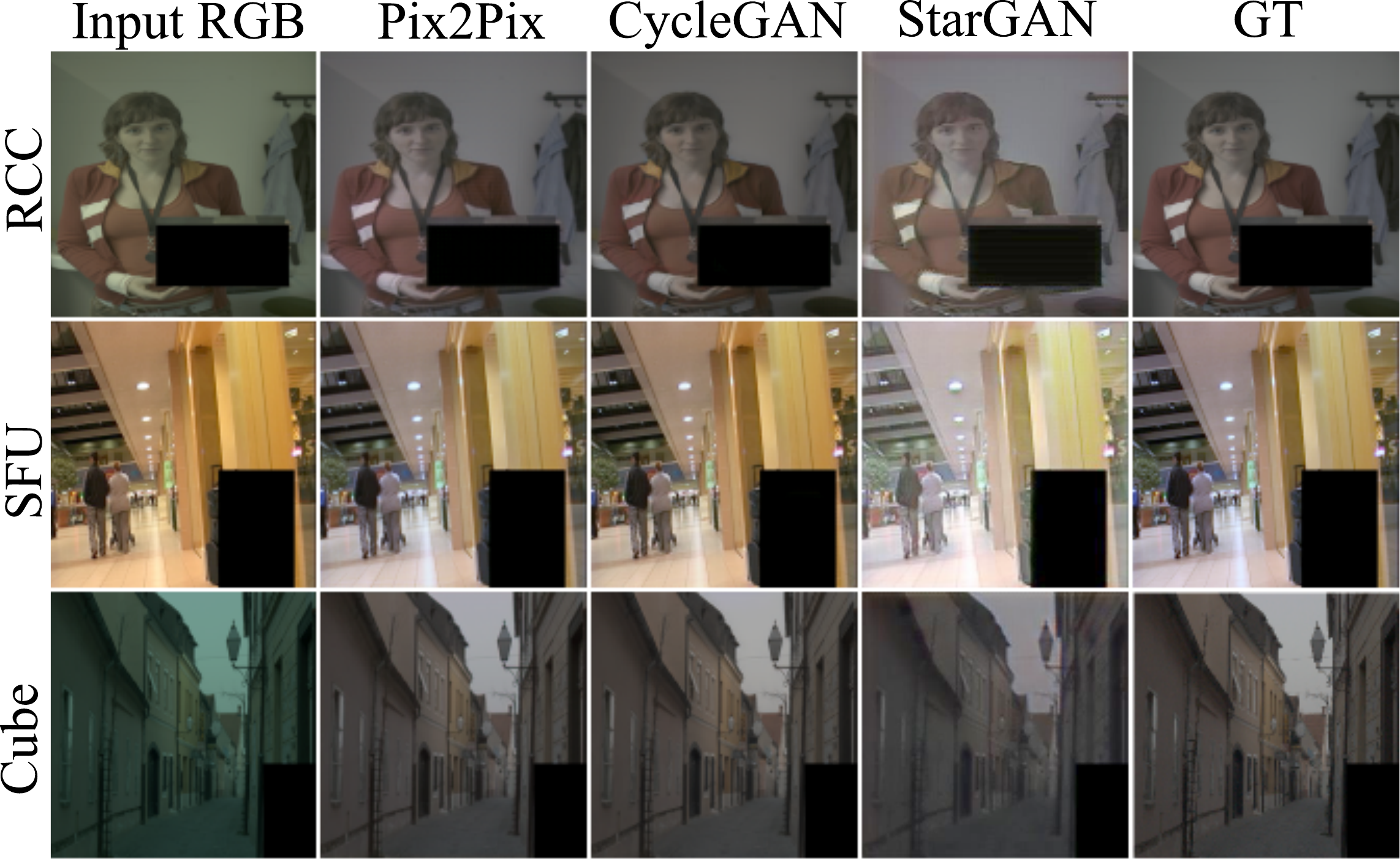}
    \caption{Performance of different GAN models on the color constancy task. All models can recover proper white balanced images. The images are gamma adjusted for visualization.}
    \label{fig:architectures}
\end{figure}

In this experiment, the performance of 3 different state-of-the-art GAN models for color constancy is given: Pix2Pix~\cite{Isola2017}, CycleGAN~\cite{Zhu2017} and StarGAN~\cite{Choi2018}. All models use $sRGB$ inputs and generate a white balanced image. The color of the light source is estimated by Equation~\eqref{eq:metric}. The results are presented in~\cref{tab:gan_rec,tab:gan_cube,tab:gan_ball} for the different datasets. Figure~\ref{fig:architectures} shows some sample outputs.

\begin{table}
\centering
\scalebox{0.75}{
\begin{tabular}{|l|c|c|c|c|c|c|}
\hline
Model & Mean & Med. & Tri. & Best 25\% & Worst 25\% & Max \\
\hline \hline
Pix2Pix~\cite{Isola2017} & \textbf{6.6} & \textbf{5.3} & \textbf{5.4} & \textbf{1.4} & \textbf{14.2} & \textbf{36.0}\\ \hline
StarGAN~\cite{Choi2018}& 10.3&8.9&9.0&4.0&18.9&46.0\\
\hline
CycleGAN~\cite{Zhu2017} & 8.4 & 5.9 & 6.4 & 1.5 & 19.6 & 37.8\\ 
\hline
\end{tabular}}
\vspace{+1mm}
\caption{Performance of different GAN models for SFU Gray Ball dataset~\cite{Ciurea2003}. Pix2Pix achieves the best performance.}\label{tab:gan_ball}
\end{table}

\begin{table}
\centering
\scalebox{0.75}{
\begin{tabular}{|l|c|c|c|c|c|c|}
\hline
Model & Mean & Med. & Tri. & Best 25\% & Worst 25\% & Max \\
\hline \hline
Pix2Pix~\cite{Isola2017} & 3.6 & 2.8 & 3.1 & 1.2 & \textbf{7.2} & \textbf{11.3}\\ \hline
StarGAN~\cite{Choi2018} & 5.3&4.2&4.6&1.5&11.0&21.8\\ \hline
CycleGAN~\cite{Zhu2017} & \textbf{3.4} & \textbf{2.6} & \textbf{2.8} & \textbf{0.7} & 7.3 & 18.0\\
\hline
\end{tabular}}
\vspace{+1mm}
\caption{Performance of different GAN models for ColorChecker RECommended dataset~\cite{Finlayson2017}. CycleGAN achieves the best performance.}
\label{tab:gan_rec}
\end{table}

\begin{table}
\centering
\scalebox{0.75}{
\begin{tabular}{|l|c|c|c|c|c|c|}
\hline
Model & Mean & Med. & Tri. & Best 25\% & Worst 25\% & Max \\
\hline \hline
Pix2Pix~\cite{Isola2017} & 1.9 & 1.4 & 1.5 & 0.7 & 4.0 & 8.0\\ \hline
StarGAN~\cite{Choi2018} & 3.8&3.3&3.5&1.3&7.0&11.4\\ \hline
CycleGAN~\cite{Zhu2017} & \textbf{1.5} & \textbf{1.2} & \textbf{1.3} & \textbf{0.5} & \textbf{3.0} & \textbf{6.0}\\
\hline
\end{tabular}}
\vspace{+1mm}
\caption{Performance of different GAN models for Cube dataset~\cite{Banic2018}. CycleGAN achieves the best performance.}
\label{tab:gan_cube}
\end{table}

\cref{tab:gan_rec,tab:gan_cube} show that \emph{CycleGAN} outperforms \emph{Pix2Pix} and \emph{StarGAN}. Table~\ref{tab:gan_ball} shows that Pix2Pix achieves the best performance for the SFU Gray Ball dataset. Pix2Pix learns a per-pixel mapping of the illumination and is agnostic about the global illumination or the spatial relations. On the other hand, the other GAN models learn a global representation of the scene illumination. Since CycleGAN can learn a more global transformation, it is able to outperform the other GANs for the color Checker and Cube dataset. However, for the Gray ball dataset the ground truth estimation is prone to errors. In conclusion, the global estimate of CycleGAN fits all the light sources under a single transformation function, while Pix2Pix can work around this by learning a per-pixel independent transformation.

\begin{figure}
    \centering
    \includegraphics[width=0.9\linewidth]{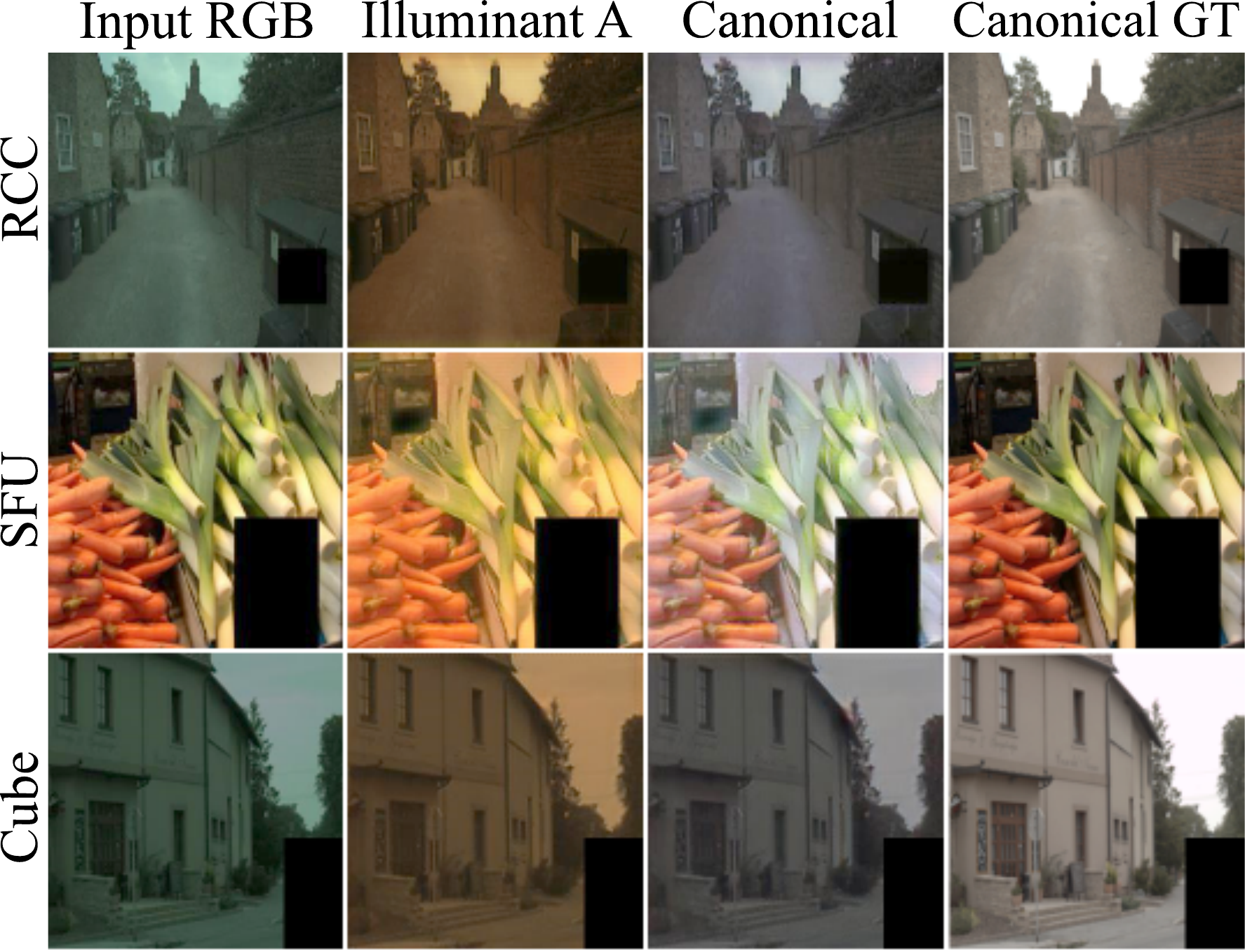}
    \caption{Visualisations of the StarGAN architecture. This architecture learns mappings to multiple domains, e.g., Illuminant A and canonical, simultaneously. The images are gamma adjusted for visualisation.}
    \label{fig:stargan}
\end{figure}

StarGAN has much lower performance than the other 2 models. This may be due to the architecture's latent representation that has to compensate for learning 3 different (input, canonical, Illuminant A) domain transformations. This is not the case with either Pix2Pix or CycleGAN, where the mapping is one-to-one. The multiple target illumination domain output of StarGAN can be beneficial, as shown in Figure~\ref{fig:stargan}, for relighting tasks. The accuracy of this internal estimation defines how close the output is to the target domain. Ideally, this estimated ground-truth illumination should be the same for all the output domains. Hence, by checking the performance of the output in estimating the ground-truth using Equation~\eqref{eq:metric}, the consistency of the network can be checked. This also allows for observing which illumination performs better in estimating the ground-truth illuminants. Per-illuminant consistency performance is presented in Table~\ref{tab:stargan_re_data}. It can be observed from the table that for the RECommended Color Checker (RCC) dataset \emph{Illuminant A} estimation yields better results than the direct canonical estimation. For the Cube dataset, the canonical performs better. The Cube dataset consists of only outdoor images. Hence, the input images seldom have an illumination that is closer to Illuminant A. Conversely, for the RCC dataset, indoor scenes generally have incandescent lighting, which is closer to Illuminant A.

\begin{table}
\centering
\scalebox{0.71}{
\begin{tabular}{|l|c|c|c|c|c|c|}
\hline
Dataset + Illumination & Mean  & Med. & Tri. & Best 25\% & Worst 25\% & Max \\
\hline \hline
RCC + Canonical & 5.3 & 4.2 & 4.6 & 1.5 & 11.0 & 21.8\\
RCC + Illuminant A & 2.9 & 2.2 & 2.3 & 0.8 & 6.4 & 13.8\\ \hline
Cube + Canonical & 3.8 & 3.3 & 3.5 & 1.3 & 7.0 & 11.4\\
Cube + Illuminant A & 4.1 & 3.2 & 3.4 & 1.1 & 8.5 & 20.7\\ \hline
SFU + Canonical & 10.3 & 8.9 & 9.0 & 4.0 & 18.9 & 46.0\\
SFU + Illuminant A & 12.7 & 11.2 & 11.2 & 3.2 & 24.8 & 61.1\\
\hline
\end{tabular}}
\vspace{+1mm}
\caption{Consistency of StarGAN's illumination estimation.}
\label{tab:stargan_re_data}
\end{table}

\subsubsection{Comparison to the State-of-the-Art}
In this section, the GAN models are compared with different benchmarking algorithms including Grey-World~\cite{Buchsbaum1980}, Second-order Grey-Edge~\cite{Weijer2007}, Pixel-based Gamut~\cite{Barnard2000}, Intersection-based Gamut~\cite{Barnard2000}, Edge-based Gamut~\cite{Barnard2000}, Spatial Correlations\cite{Chakrabarti2012}, Natural Image Statistics~\cite{Gijsenij2010}, High Level Visual Information~\cite{Weijer2007-2}, Exemplar-Based color constancy~\cite{Joze2014}, Color Tiger~\cite{Banic2018}. Furthermore, comparisons to 2 convolutional approaches are given: Deep color constancy~\cite{Bianco2015} and Fast Fourier color constancy~\cite{Barron2016}. Some of the results are taken from related papers, resulting in missing entries for some datasets.~\cref{tab:external_grayball,tab:external_cc_re,tab:external_cube} provide quantitative results for 3 different benchmarks. 

\begin{table}
\centering
\scalebox{0.68}{
\begin{tabular}{|l|c|c|c|c|c|c|}
\hline
Method & Mean  & Med. & Tri. & Best 25\% & Worst 25\% & Max \\
\hline \hline
Grey-World~\cite{Buchsbaum1980} & 13.0 & 11.0 & 11.5 & 3.1 & 26.0 & 63.0\\
Edge-based Gamut~\cite{Barnard2000} & 12.8 & 10.9 & 11.4 & 3.6 & 25.0 & 58.3\\
Spatial Correlations~\cite{Chakrabarti2012} & 12.7 & 10.8 & 11.5 & 2.4 & 26.1 & 41.2\\
Second-order Grey-Edge~\cite{Weijer2007} & 10.7 & 9.0 & 9.4 & 3.2 & 20.9 & 56.0\\
Bottom-up \& Top-down~\cite{Weijer2007-2} & 9.7 & 7.7 & 8.2 & 2.3 & 20.6 & 60.0\\
Natural Image Statistics~\cite{Gijsenij2010} & 9.9 & 7.7 & 8.3 & 2.4 & 20.8 & 56.1\\
E. B. color constancy~\cite{Joze2014} & 8.0 & 6.5 & 6.8 & 2.0 & 16.6 &53.6\\
\hline
Pix2Pix~\cite{Isola2017} & \textbf{6.6} & \textbf{5.3} & \textbf{5.4} & \textbf{1.4} & \textbf{14.2} & \textbf{36.0}\\
CycleGAN~\cite{Zhu2017} & 8.4 & 5.9 & 6.4 & 1.5 & 19.6 & 37.8\\
StarGAN~\cite{Choi2018} & 11.4 & 9.2 & 10.0 & 3.8 & 21.8 & 41.4\\
\hline
\end{tabular}}
\vspace{+1mm}
\caption{Performance on SFU Gray Ball~\cite{Ciurea2003}.}
\label{tab:external_grayball}
\end{table}

\begin{table}
\centering
\scalebox{0.68}{
\begin{tabular}{|l|c|c|c|c|c|c|}
\hline
Method & Mean  & Med. & Tri. & Best 25\% & Worst 25\% & Max \\
\hline \hline
Grey-World~\cite{Buchsbaum1980} & 9.7 & 10.0 & 10.0 & 5.0 & 13.7 & 24.8\\
White-Patch~\cite{Land1971} & 9.1 & 6.7 & 7.8 & 2.2 & 18.9 & 43.0\\
Shades-of-Grey~\cite{Finlayson2004} & 7.3 & 6.8 & 6.9 & 2.3 & 12.8 & 22.5\\
AlexNet + SVR~\cite{Bianco2015} & 7.0 & 5.3 & 5.7 & 2.9 & 14.0 & 29.1\\
Pixel-based Gamut~\cite{Barnard2000} & 6.0 & 4.4 & 4.9 & 1.7 & 12.9 & 25.3\\
Intersection-based Gamut~\cite{Barnard2000} & 6.0 & 4.4 & 4.9 & 1.7 & 12.8 & 26.3\\
Edge-based Gamut~\cite{Barnard2000} & 5.5 & 3.3 & 3.9 & 0.7 & 13.8 & 29.8\\
Deep Colour Constancy~\cite{Bianco2015} & 4.6 & 3.9 & 4.2 & 2.3 & 7.9 & 14.8\\
Second-order Grey-Edge~\cite{Weijer2007} & 4.1 & 3.6 & 3.8 & 1.5 & 8.5 & 16.9\\
First-order Grey-Edge~\cite{Weijer2007} & 4.0 & 3.1 & 3.3 & 1.4 & 8.4 & 20.6\\
FFCC (model Q)~\cite{Barron2016} & \textbf{2.0} & \textbf{1.1} & \textbf{1.4} & \textbf{0.3} & \textbf{4.6} & 25.0\\
\hline
Pix2Pix~\cite{Isola2017} & 3.6 & 2.8 & 3.1 & 1.2 & 7.2 & \textbf{11.3}\\
CycleGAN~\cite{Zhu2017} & 3.4 & 2.6 & 2.8 & 0.7 & 7.3 & 18.0\\
StarGAN~\cite{Choi2018} & 5.7 & 4.9 & 5.2 & 1.7 & 10.5 & 19.5\\
\hline
\end{tabular}}
\vspace{+1mm}
\caption{Performance on ColorChecker RECommended~\cite{Finlayson2017}.}
\label{tab:external_cc_re}
\end{table}

\begin{table}
\centering
\scalebox{0.75}{
\begin{tabular}{|l|c|c|c|c|c|c|}
\hline
Method & Mean  & Med. & Tri. & Best 25\% & Worst 25\% \\
\hline \hline
Grey-World~\cite{Buchsbaum1980} & 3.8 & 2.9 & 3.2 & 0.7 & 8.2\\
White-Patch~\cite{Land1971} & 6.6 & 4.5 & 5.3 & 1.2 & 15.2\\
Shades-of-Grey~\cite{Finlayson2004} & 2.6 & 1.8 & 2.9 & 0.4 & 6.2\\
General Grey-World~\cite{Barnard2002} & 2.5 & 1.6 & 1.8 & 0.4 & 6.2\\
Second-order Grey-Edge~\cite{Weijer2007} & 2.5 & 1.6 & 1.8 & 0.5 & 6.0\\
First-order Grey-Edge~\cite{Weijer2007} & 2.5 & 1.6 & 1.8 & 0.5 & 5.9\\
Color Tiger~\cite{Banic2018} & 3.0 & 2.6 & 2.7 & 0.6 & 5.9\\
Restricted Color Tiger~\cite{Banic2018} & 1.6 & \textbf{0.8} & \textbf{1.0} & \textbf{0.2} & 4.4\\
\hline
Pix2Pix~\cite{Isola2017} & 1.9 & 1.4 & 1.5 & 0.7 & 4.0\\
CycleGAN~\cite{Zhu2017} & \textbf{1.5} & 1.2 & 1.3 & 0.5 & \textbf{3.0}\\
StarGAN~\cite{Choi2018} & 4.8 & 3.9 & 4.2 & 1.9 & 8.9\\
\hline
\end{tabular}}
\vspace{+1mm}
\caption{Performance on Cube dataset\cite{Banic2018}.}
\label{tab:external_cube}
\end{table}

It is shown in Table~\ref{tab:external_grayball} (SFU Gray Ball), Pix2Pix~\cite{Isola2017} achieves the best performance in all the metrics with 17.5\% improvement in mean angular error, 18.4\% in median and 20.6\% in trimean. For Table~\ref{tab:external_cc_re} (RCC), CycleGAN~\cite{Zhu2017} framework achieves the second best place in different metrics, only worse than Fast Fourier color constancy~\cite{Barron2016}. In Table~\ref{tab:external_cube} (Cube), CycleGAN~\cite{Zhu2017} achieves the best performance in the mean error and worst and comparable results with all other methods. 

\subsubsection{GAN Performance for Multiple Light Sources}

In this section, the performance of GANs in a multi-illumination setting is studied. The models are trained on 3 datasets and tested on the MIMO dataset. The results are shown in~\cref{tab:pix2pix_mimo,tab:cycle_mimo}. The performance of StarGAN is not provided since, from the experiments, it is observed: 1) to be generally the worst performing, and 2) CycleGAN is a simplified version of StarGAN with just one domain transform.

\begin{table}
\centering
\resizebox{0.2\textwidth}{!}{
\begin{tabular}{|c|c|c|}
\hline
Dataset & Mean & Median \\ \hline
Cube    & 18.7 & 18.1   \\ \hline
REC-CC  & 14.8 & 14.5   \\ \hline
SFU     & 20.1 & 17.8   \\ \hline
\end{tabular}%
}
\caption{Performance of Pix2Pix on the MIMO dataset.}
\label{tab:pix2pix_mimo}
\end{table}

\begin{table}
\centering
\resizebox{0.2\textwidth}{!}{
\begin{tabular}{|c|c|c|}
\hline
Dataset & Mean & Median \\ \hline
Cube    & 12.7 & 11.9   \\ \hline
REC-CC  & 13.3 & 12.9   \\ \hline
SFU     & 20.1 & 17.4   \\ \hline
\end{tabular}%
}
\caption{Performance of CycleGAN on the MIMO dataset.}
\label{tab:cycle_mimo}
\end{table}

The results show that both the algorithms are unable to cope with images with multiple light sources. This is because both models are trained to learn a single illumination transformation. As a result, the models confuse object colors with illuminants. Pix2Pix shows slightly larger errors because no contextual constraints are imposed on the pixels. As such, ambiguous points like edges and boundaries contribute to higher errors. Inclusion of an explicit illumination context becomes a necessity for the network to be able to model a multi-illumination setting.

\subsection{Multi Illumination Color Constancy}

In this section, the proposed method for multi-illumination color constancy is analysed. For all the experiments, four illuminants are chosen, $N=4$. However, the proposed method is not dependent on the choice of $N$ and can be arbitrarily scaled. All corresponding hyper-parameters of the model are kept constant. This is to ensure that the influence of the components being tested is not influenced by a change in the hyper-parameters. For training, the entire set of 5000 images from Intel TUT is used, while testing is done exclusively on the MIMO dataset.

\subsubsection{Influence of Adversarial Loss}

In this experiment, the influence of the adversarial loss is studied. Since there is a direct supervision from the ground truth, the network is trained in the traditional manner, where an encoder-decoder architecture is used to produce the output. Both the generator and the discriminator losses are removed and the network is trained using only the L1 losses on the predicted illumination map~\eqref{eq:loss_illum}, the color corrected image~\eqref{eq:loss_rgb} and the individual seed consistency~\eqref{eq:loss_masks}. The results for the experiment are shown in table~\ref{tab:gan_less}.

\begin{table}
\centering
\resizebox{0.25\textwidth}{!}{
\label{tab:gan_less}
\begin{tabular}{|c|c|c|}
\hline
Algorithm    & Mean & Median \\ \hline
Without Adv. Loss & 3.7  & 3.5    \\ \hline
Proposed     & \textbf{3.5}  & \textbf{2.9}    \\ \hline
\end{tabular}%
}
\caption{Influence of the GAN loss.}
\end{table}

It is observed that omitting the adversarial loss reduces the performance. Adding the adversarial loss helps the model to cope with outliers, while also improving the average performance of the model. This can be explained as follows. The discriminator in a GAN is actively seeking for unrealistic outputs, which are often outliers. Furthermore, without the adversarial losses, the network only minimizes a distance function, which can lead to blurry outputs or ambiguous edges and borders. Since the proposed method is pixel-wise, these errors propagate more strongly through the network. In conclusion, this experiment shows that the inclusion of the adversarial loss is a necessary component for the proposed formulation.

\subsubsection{Influence of Illumination Probability Map Prediction}

\begin{figure}
    \centering
    \includegraphics[width=0.7\linewidth,height=0.3\textwidth]{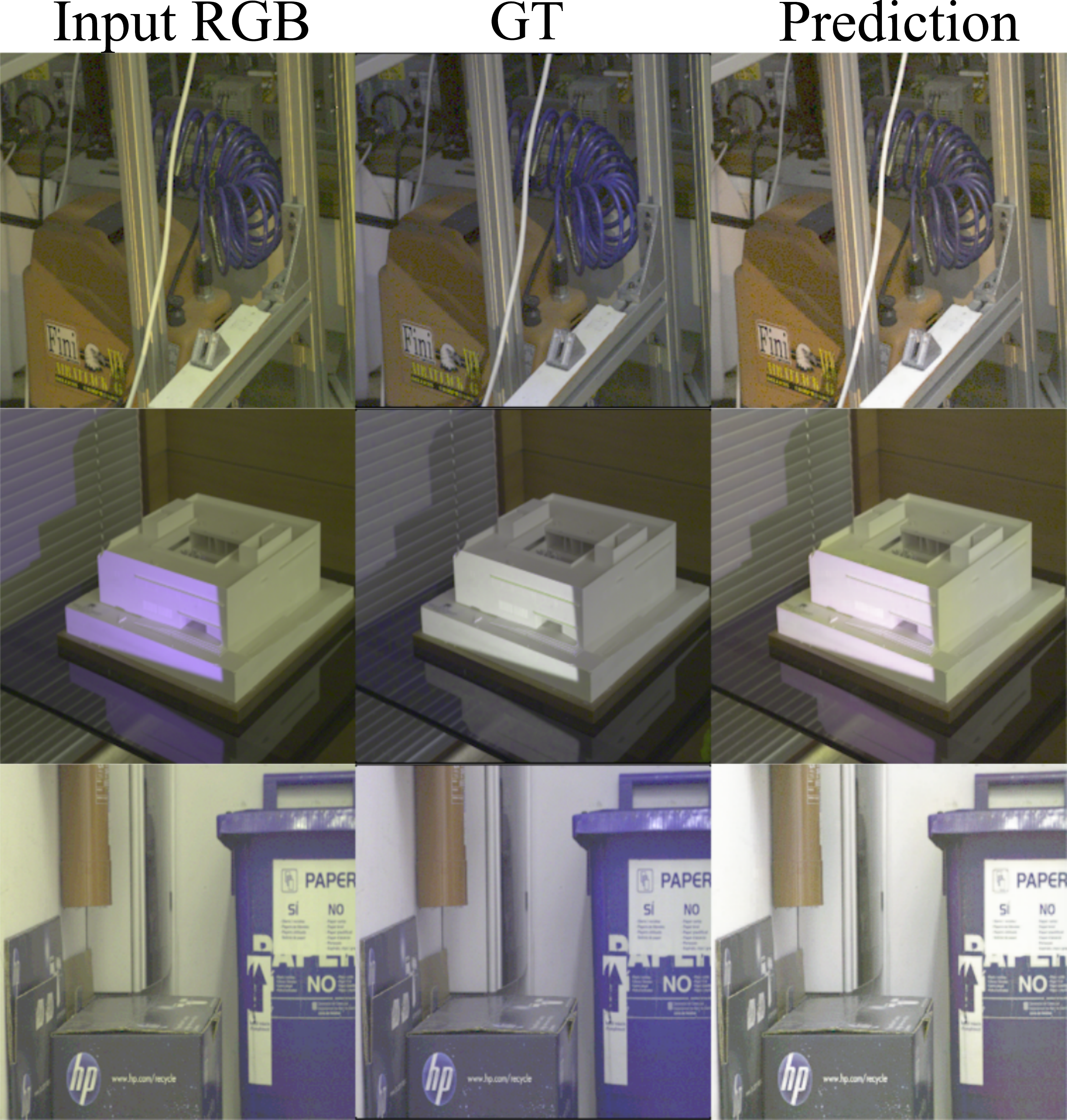}
    \caption{Comparison of the modified direct correction model to the ground truth. Here, the model directly outputs the color corrected image, instead of producing an illumination probability map. It is observed that the model is not able to fully correct for the illumination and as such, has residual illumination colors (2nd row, the printer.) This can be explained by the lack of the explicit probability map, which forces the network to instead predict a uniform illumination.}
    \label{fig:no_prob_visuals}
\end{figure}

In the previous experiment, we modelled the problem as an image-to-image translation. In this experiment, we study such a formulation for the proposed method. For this, instead of predicting a per-pixel probability map, the color corrected image is directly predicted. For the illumination losses, the predicted illumination map is obtained by using the predicted color corrected image and the input color bias image itself. This experiment verifies whether adding the samples as an extra input to Pix2Pix improves the performance. It is also shown to what extent Pix2Pix can adapt to a multi-illumination setting. The results are given in table~\ref{tab:prob_less}. Figure~\ref{fig:no_prob_visuals} shows some outputs.

\begin{table}
\centering
\resizebox{0.3\textwidth}{!}{
\label{tab:prob_less}
\begin{tabular}{|c|c|c|}
\hline
Algorithm                  & Mean  & Median \\ \hline
Without Probability Output & 5.30 & 5.24  \\ \hline
Proposed                   & \textbf{3.5}   & \textbf{2.9}    \\ \hline
\end{tabular}%
}
\caption{Influence of a probability map as the output.}
\end{table}

The results show that regressing the color corrected image directly using the proposed formulation yields higher errors. Since the model takes initial seed points and tries to constrain the output to be consistent with the seed points, the model fails to perform a full correction. This is because with a probability output, the initial seed points are directly linked to the illuminants. However, having a color corrected output needs the network to learn this probability implicitly. This, coupled with the illumination constraints on the loss functions, causes the network to leak the illumination over to the image directly. From the figure it can also be observed that the outputs are a bit more washed out and contain illumination color as residuals. This is because without the probability map support, the network compensates by averaging the illumination distribution more uniformly.

\subsubsection{Comparison to State of the Art}

\begin{figure}
    \centering
    \includegraphics[width=0.8\linewidth]{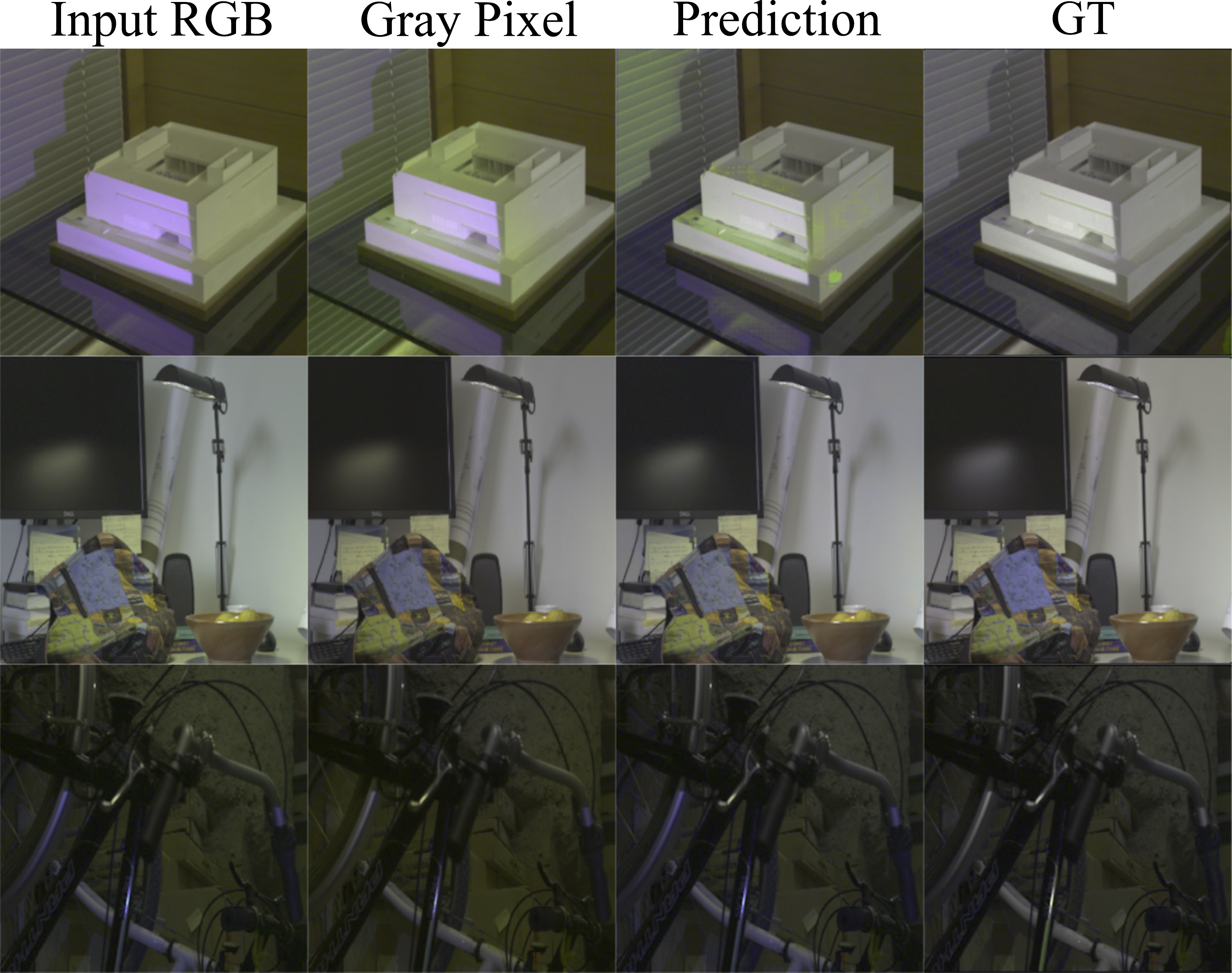}
    \caption{Comparison of the proposed method to sota methods. The proposed method can properly recover color corrected images. Gray Pixel can somewhat handle multi-illumination. However, it fails for ambiguous cases like the bike (3rd row). Gray Pixel fails to remove the yellow color cast of the light. The proposed method can recover such ambiguities.}
    \label{fig:sota_visuals}
\end{figure}

In this experiment, the performance of the proposed method is compared with the state-of-the art methods. The results are shown in table~\ref{tab:sota_results}. A visual comparison is also provided in figure~\ref{fig:sota_visuals}. The Grayness Index algorithm is compared against 4-illumination clustering since the proposed method is tested on the 4-illumination assumption. A multi-illumination setting, like indoors, is more likely to have $4$ or less major illuminants. Hence $N$ is set to $4$ for the experiments.

\begin{table}
\centering
\resizebox{0.2\textwidth}{!}{
\begin{tabular}{|c|c|c|}
\hline
Algorithm     & Mean & Median \\ \hline
Doing Nothing & 8.9  & 8.8    \\ \hline
~\cite{Gijsenij2012}      & 3.8  & 4.2    \\ \hline
CRF~(\cite{Beigpour2014})           & 4.1  & 3.3    \\ \hline
GI(M=4)~(\cite{Qian2019}       & 4.0  & 3.5    \\ \hline
Proposed      & \textbf{3.5}  & \textbf{2.9}    \\ \hline
\end{tabular}%
}
\caption{Comparison to State-of-the-Art algorithms. }
\label{tab:sota_results}
\end{table}

It is shown that the proposed model can outperform all the baselines by a comfortable margin. From the figure, it is observed that the proposed method can properly recover color corrected images. The proposed method can correct for the illumination on the printer, while Grayness Index classifies that as part of the object color. Similarly, Grayness Index is unable to color correct the reflection on the monitor while the proposed method can. The only requirement for the proposed model is the corresponding seed points for each illumination, only required for the training phase. This requirement can be easily satisfied as discussed previously using a single illumination dataset and augmenting it, without requiring any specialised hardware. For edge cases of more illuminants in the test image, compared to the $N$ illumination number during the training, the network will simply collapse them into $N$ illuminants. Explicit handling of such cases is beyond the scope of this work and can be a future research.

\section{Conclusion}

In this paper, we modelled the problem of color constancy as an image-to-image translation. GANs are used to learn domain transfers. A hybrid approach has been proposed. A single-illumination dataset augmentation has also been proposed to address the problem of the lack of large multi-illumination datasets to enable learning. Experiments on both single and multi-illumination datasets show that our methods outperform sota methods.

\title{Supplementary Material}
\suptit

\section{Network Details}
\begin{figure}[ht]
    \centering
    \includegraphics[width=0.8\linewidth]{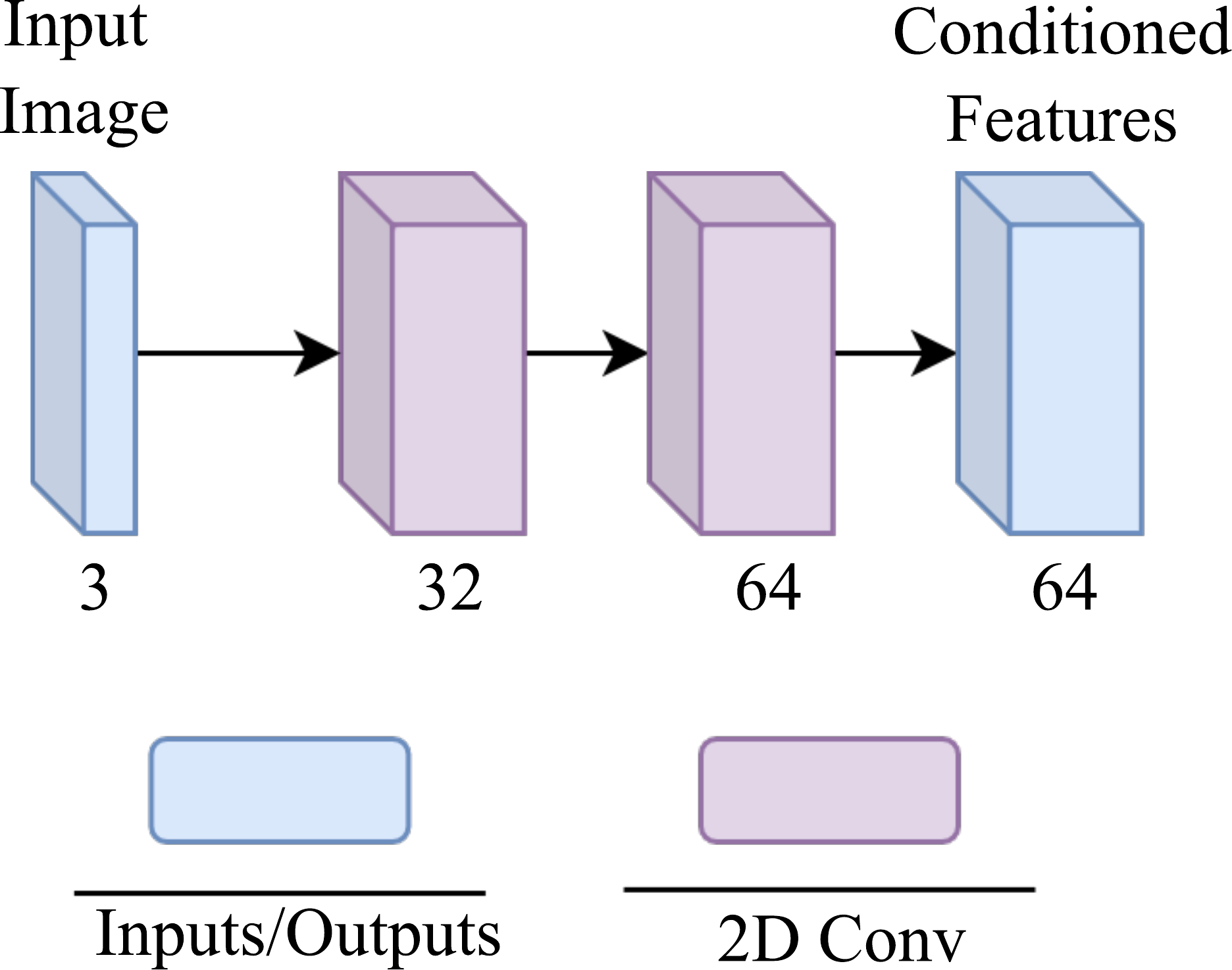}
    \caption{The architecture for the image processing head is shown. The input image is passed through two convolution layers, which squeezes and subsequently expands the channels of the outputs, thus conditioning the features down to the useful features in an end-to-end fashion.}
    \label{fig:image_head}
\end{figure}

\begin{figure}[ht]
    \centering
    \includegraphics[width=0.8\linewidth]{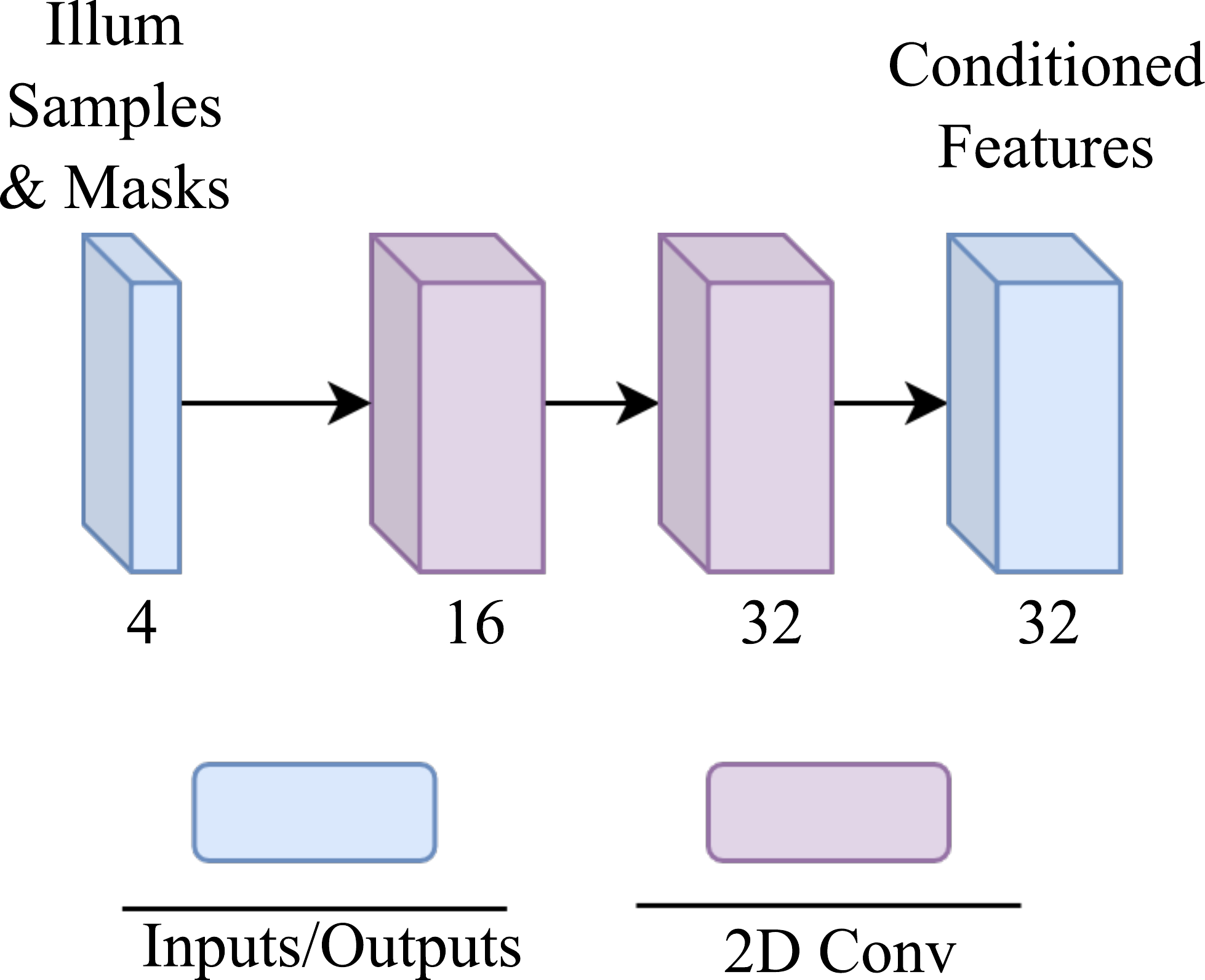}
    \caption{The architecture for the illumination processing head is shown. The input is the sparse illumination samples and the corresponding masks, concatenated channel wise ($3 + 1$ channels). This is passed through two convolution layers, which squeezes and subsequently expands the channels of the outputs, thus conditioning the features down to the useful features in an end-to-end manner.}
    \label{fig:illum_head}
\end{figure}

The architectural details are provided. The inputs to the network are first processed through a shallow convolution head, for feature conditioning. The image processing  and the illumination processing head are detailed separately in~\ref{fig:image_head} and~\ref{fig:illum_head}, respectively. The input to the image processing head is the $RGB$ color biased image, while the illumination processing head takes the concatenation of the sparse illumination map and the mask as input. The output for the image head and illumination head is a 64 channel and 32 channel tensor, respectively. All of them have the same spatial dimensions as the input.~\cref{tab:head_branch,tab:illum_branch} details the configurations.

\begin{table}[ht]
\centering
\resizebox{0.48\textwidth}{!}{%
\begin{tabular}{|l|l|l|l|}
\hline
Name       & \textbf{Layer} & \textbf{Kernel Size, Stride, Padding} & \textbf{Output Size} \\ \hline
Input      & conv1 & 3x3x32, 1, 1                 & 256x256x32  \\ \hline
Output     & conv2 & 3x3x32, 1, 1                 & 256x256x64  \\ \hline
\end{tabular}%
}
\caption{Overview of the image processing head. The input is 3 channel and the output is a 64 channel tensor with the same spatial dimension.}
\label{tab:head_branch}
\end{table}

\begin{table}[ht]
\centering
\resizebox{0.48\textwidth}{!}{%
\begin{tabular}{|l|l|l|l|}
\hline
Name       & \textbf{Layer} & \textbf{Kernel Size, Stride, Padding} & \textbf{Output Size} \\ \hline
Input      & conv1 & 3x3x16, 1, 1                 & 256x256x16  \\ \hline
Output     & conv2 & 3x3x16, 1, 1                 & 256x256x32  \\ \hline
\end{tabular}%
}
\caption{Overview of the illumination processing head. The input is 4 channel concatenation of the sparsse illumination samples and illumination masks. The output is a 32 channel tensor with the same spatial dimension.}
\label{tab:illum_branch}
\end{table}

The Pix2Pix~\cite{Isola2017} generator architecture is used, with just two changes: 1) The input channels are changed to match the incoming tensors' dimension from the processing heads and, 2) The final activation is changed to Sigmoid instead of Tanh, to output a probability map instead of an image.

\clearpage

{\small
\bibliographystyle{ieee_fullname}
\bibliography{egbib}
}

\end{document}